\documentclass[9pt,twocolumn,twoside]{opticajnl}
\journal{opticajournal} 

\setboolean{shortarticle}{false}

\usepackage{lineno}
\usepackage{color}

\newcommand{\add}[1]{\textcolor{black}{#1}}

\title{Neural radiance fields-based holography [Invited]}

\author[1]{Minsung Kang}
\author[1]{Fan Wang}
\author[1]{Kai Kumano}
\author[1]{Tomoyoshi Ito}
\author[1,*]{Tomoyoshi Shimobaba}

\affil[1]{Graduate School of Engineering, Chiba University, 1-33, Yayoi-cho, Chiba-shi, Inage-ku, Chiba 263-8522, Japan}

\affil[*]{shimobaba@faculty.chiba-u.jp}

\begin{abstract}
This study presents a novel approach for generating holograms based on the neural radiance fields (NeRF) technique. \add{Generating real-world three-dimensional (3D)} data is difficult in hologram computation. NeRF is a state-of-the-art technique for 3D light-field reconstruction from 2D images based on volume rendering.
\add{The NeRF can rapidly predict new-view images that are not included in a training dataset.}
In this study, we constructed a rendering pipeline directly from \add{a radiance field} generated from 2D images by NeRF for hologram generation using deep neural networks within a reasonable time. 
The pipeline comprises three main components: the NeRF, a depth predictor, and a hologram generator, all constructed using deep neural networks. 
The pipeline does not include any physical calculations.
The predicted holograms of a 3D scene viewed from any direction were computed using the proposed pipeline.
The simulation and experimental results are presented.
\end{abstract}

\setboolean{displaycopyright}{false} 

\begin{document}
\maketitle

\section{Introduction}
Holographic displays are promising three-dimensional (3D) displays that have been actively researched in recent years because they fulfill all the human factors necessary for stereoscopic viewing without discomfort \cite{hilaire1992synthetic,poon2006digital}.
Holographic displays necessitate a three-step process: first, generate 3D scene data using computer graphics techniques or 3D cameras, such as time-of-flight cameras and light-field cameras; second, produce holograms from the 3D scene data; and finally, reproduce 3D images from the generated holograms.
However, each step presents obstacles to its practical implementation.

Spatial light modulators (SLMs) for displaying holograms inherently require extremely large spatial bandwidth products (SBPs) to simultaneously achieve a large field of view and a wide viewing angle for 3D reconstruction \cite{blinder2019signal}. 
Future holographic displays require giga-to-tera-pixel SLMs, and currently, no commercial SLMs satisfy the requirements of SBPs.
To address this challenge, spatial and temporal methods using conventional SLMs to enhance SBPs have been proposed \cite{takaki2014viewing,sasaki2014large,li2018full,park2019ultrathin}. These systems have successfully reconstructed 3D scenes with a large field-of-view and wide viewing angles. 
Special materials with large SPB, such as updatable photorefractive polymers \cite{tay2008updatable,blanche2010holographic} and magneto-optical materials \cite{takagi2014magneto,makowski2022dynamic}, have also garnered attention.
However, an increase in SBP requirements will lead to further increases in the computational costs associated with generating 3D scene data and holograms.

The calculation of holograms and generation of 3D scenes are also obstacles.
Holograms are computed based on a light propagation model and can be categorized into point clouds \cite{yamaguchi2007real}, polygons \cite{matsushima2009extremely,wang2023high}, light fields \cite{wakunami2011calculation,zhao2015accurate,yamaguchi2016light}, and deep-learning approaches. 
The first two represent the 3D scene either as a collection of point cloud data or polygon data. They utilize a point spread function or transfer function representation to calculate light waves from these primitives.
\add{Conversely, light-field and layered hologram approaches show promise in hologram computation.} These approaches automatically render illumination, shading, and hidden surface removal, which are typically cumbersome in point-cloud and polygon approaches. However, this method tends to sacrifice the resolution of the 3D reconstruction.
Deep learning has recently emerged as a new hologram-computation technique \cite{horisaki2018deep,goi2020deep,eybposh2020deepcgh,shi2021towards,choi2021neural,shi2022end}.
These studies have revealed excellent computational performance and image quality; however, all the aforementioned studies require cumbersome and time-consuming 3D scene generation.
Several studies proposed the generation of 3D holograms without 3D scene data \cite{chang2023picture,ishii2023multi}. 
These approaches do not require special 3D cameras or computer graphics.
These neural networks can predict 3D holograms from 2D images. However, the predicted holograms are only generated from a single viewpoint. In addition, training these neural networks requires a diverse and large 3D dataset for arbitrary 3D reconstruction.

\begin{figure*}[h]
\centering
\fbox{\includegraphics[width=\linewidth]{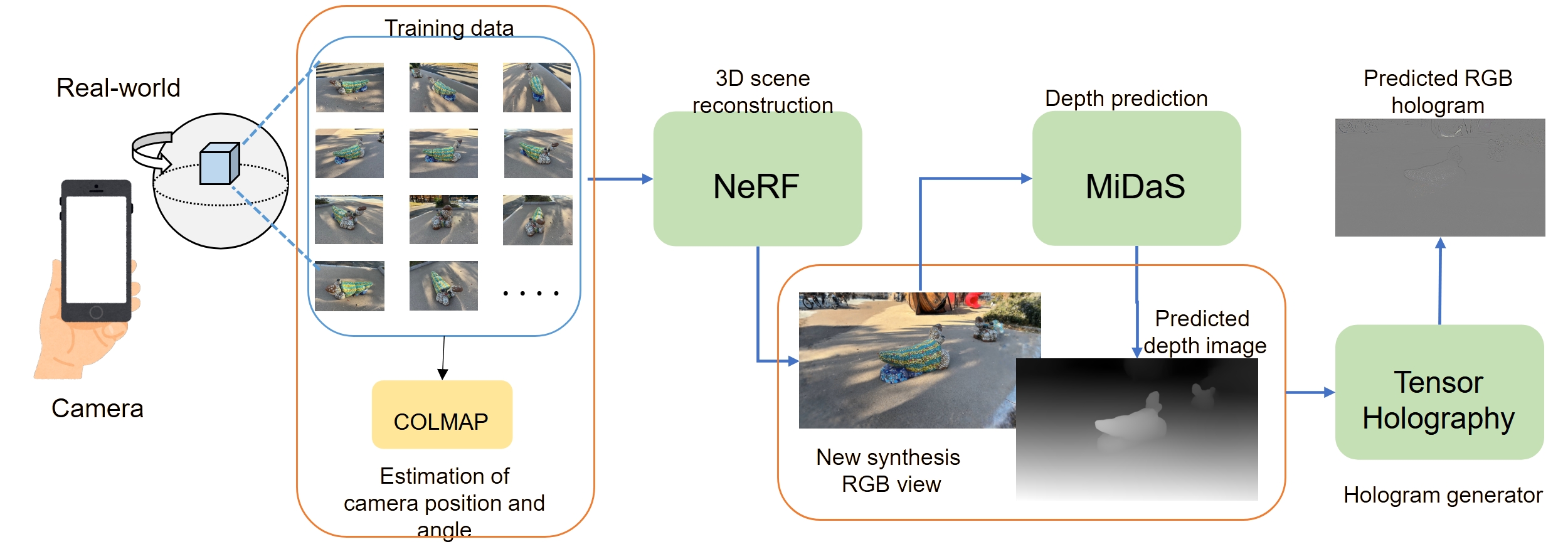}}
\caption{3D hologram generating pipeline. The first stage performs NeRF to reconstruct a \add{3D radiance field} from 2D images. The second stage predicts depth maps using MiDaS, and the last stage predicts holograms using Tensor holography.}
\label{fig:outline}
\end{figure*}

This study proposes a neural radiance fields (NeRF)-based hologram generation method that can predict holograms directly from new synthesis views without using 3D cameras or 3D graphics processing pipelines. 
The generation of 3D data is a complicated process in hologram generation. 
NeRF is a state-of-the-art technique for 3D light-field reconstruction from 2D images \cite{mildenhall2021nerf} that rapidly generates new views that are not included in the training dataset.
This study constructs a hologram-rendering pipeline directly converting from \add{a 3D radiance field} generated from 2D images using NeRFs to holograms using deep neural networks, and the pipeline predicts the final holograms within a reasonable time. 
While conventional hologram calculations involve time-consuming physical calculations, such as diffraction calculations, the pipeline does not include any physical calculations.

The remainder of this paper is organized as follows. Section 2 describes the proposed pipeline. Section 3 presents the simulation and experimental results. Section 4 discusses the advantages and limitations of the proposed pipeline, and the last section concludes the paper.

\section{Proposed pipeline}
Figure \ref{fig:outline}
illustrates the outline of the proposed pipeline.
NeRF outputs a new RGB synthesis view by inputting viewing information of the position $(x,y,z)$ and direction angles $(\phi, \theta)$; subsequently, the synthesized image is fed into a deep neural network, MiDaS \cite{ranftl2020towards}, to predict the corresponding depth map. 
Concatenating the synthesized RGB image and predicted depth image, it is input to a third deep neural network, ``Tensor Holography'' \cite{shi2021towards}, to predict a final hologram.
\add{We have chosen these neural networks that provide state-of-the-art results within a reasonable time.}

Several 3D scene reconstruction techniques have been proposed, including image-based rendering algorithms \cite{chan2007image} and structure-from-motion techniques \cite{ozyecsil2017survey}.
Among these, NeRF has advantages in terms of generation speed, quality, and novel view-synthesis abilities. 
NeRF transforms a five-dimensional vector ${\bold v}=(x, y, z, \theta, \phi)$ into an RGB image through regression using a multilayer perceptron (${\rm MLP}$) with a volume-rendering technique.
The MLP can be expressed as:
\begin{equation}
(r,g,b,\sigma) = {\rm MLP}\{{\gamma(\bold v)}\},
\end{equation}
where $\sigma$ and $(r,g,b)$ indicate the volume density and emitted radiance along the ray of ${\bold v}$.
$\gamma(x)$ denotes the positional encoder \cite{vaswani2017attention} used to represent the higher-frequency components in the 3D reconstruction.
Differentiable volume rendering $\mathcal{V}\{\cdot \}$ generates a novel view image by using
\begin{equation}
RGB = \mathcal{N}_{NeRF}\{{\bold v} \} = \mathcal{V} \left\{ {\rm MLP}\{\gamma({\bold v})\} \right\}.
\end{equation}
The loss function for training the NeRF is expressed as:
\begin{equation}
\mathcal{L} = \lVert RGB - G \rVert_2^2, 
\end{equation}
where $\lVert \cdot \rVert_2$ denotes $L_2$ norm, and $G$ denotes the ground-truth images.
In this paper, the loss function is presented in a simplified form. \add{Refer to} \cite{mildenhall2021nerf} for the actual loss function.
This study uses Instant NeRF \cite{ranftl2020towards} to accelerate the predictions.
In the training, to estimate $(x, y, z, \theta, \phi)$ for each captured image, we used COLMAP software \cite{schoenberger2016sfm}.
The captured images and their estimated $(x, y, z, \theta, \phi)$ values were fed into the MLP.
Following training, the NeRF can predict novel view images according to the new ${\bold v}$.
\begin{figure}[t]
\centering
\fbox{\includegraphics[width=\linewidth]{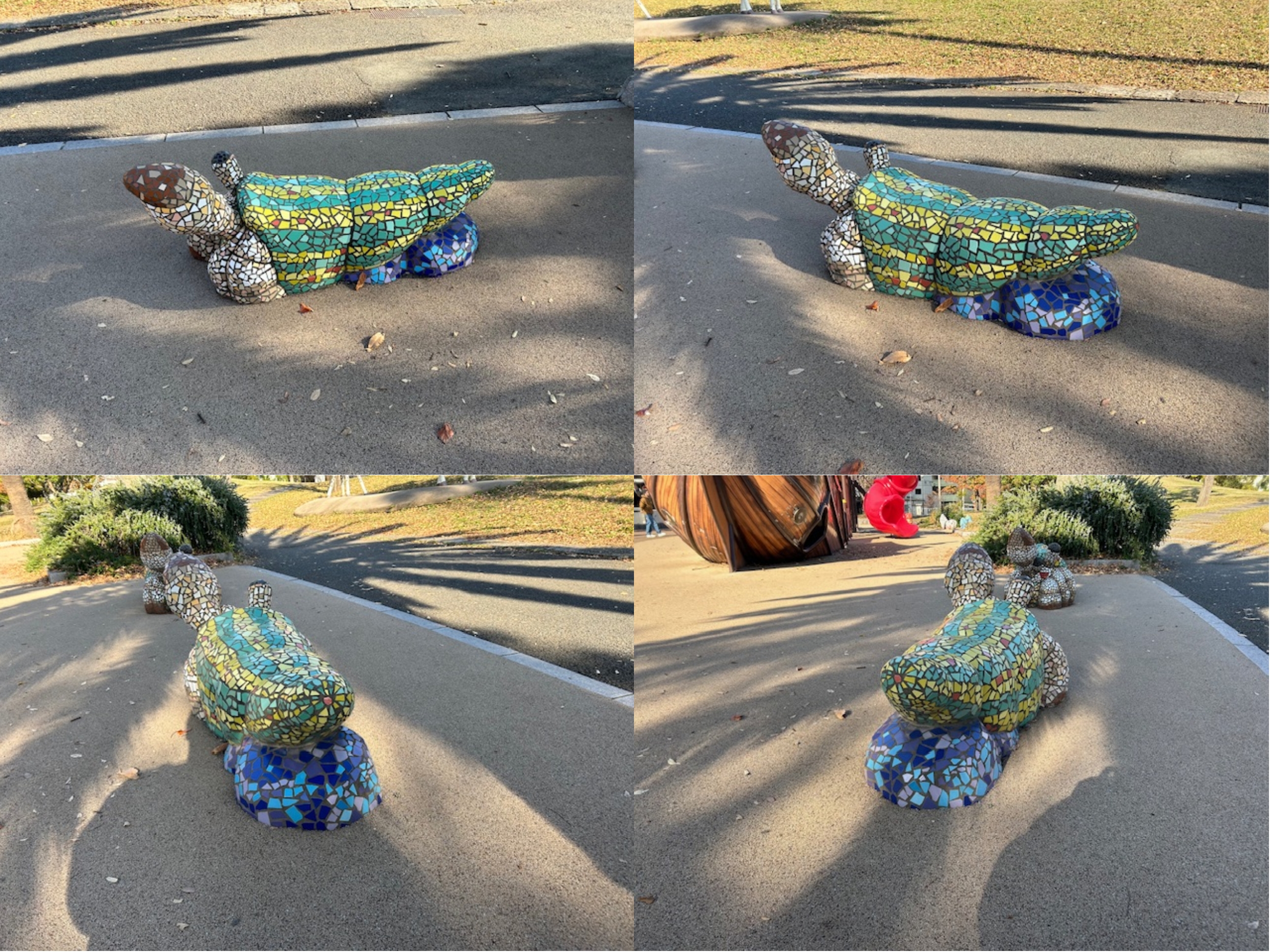}}
\caption{Park scene (hermit crab playground equipment in the center) captured by iPhone 14. The number of the photos is 70.}
\label{fig:nerf_original}
\end{figure}

\begin{figure}[h]
\centering
\fbox{\includegraphics[width=\linewidth]{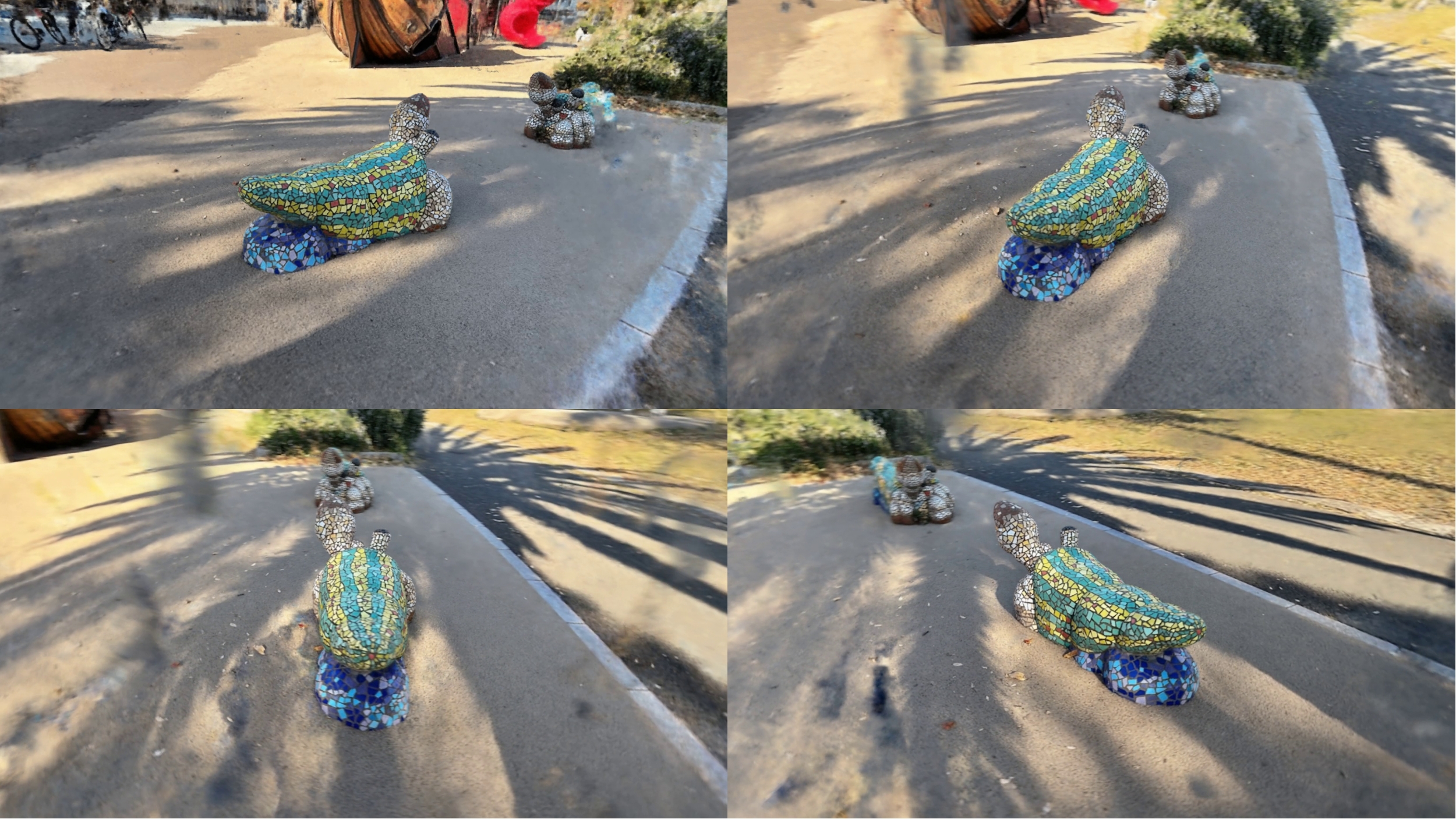}}
\caption{3D scene reconstruction from NeRF. Note that these views are not included in the original 70 images.}
\label{fig:nerf_predict}
\end{figure}

Figure \ref{fig:nerf_original} illustrates an example of captured 2D images with $640 \times 480$ pixels (ground-truth images $G$) using the iPhone 14.
The number of images used was 70.
Inputting the new parameters $(x, y, z, \theta, \phi)$ into $\mathcal{V}\left\{ {\rm MLP} \{\gamma({\bold v})\} \right\}$ produces a novel RGB image.
Figure \ref{fig:nerf_predict} illustrates an example of a predicted RGB image.
Note that these viewing images are not included in the original 70 images.

\begin{figure}[h]
\centering
\fbox{\includegraphics[width=\linewidth]{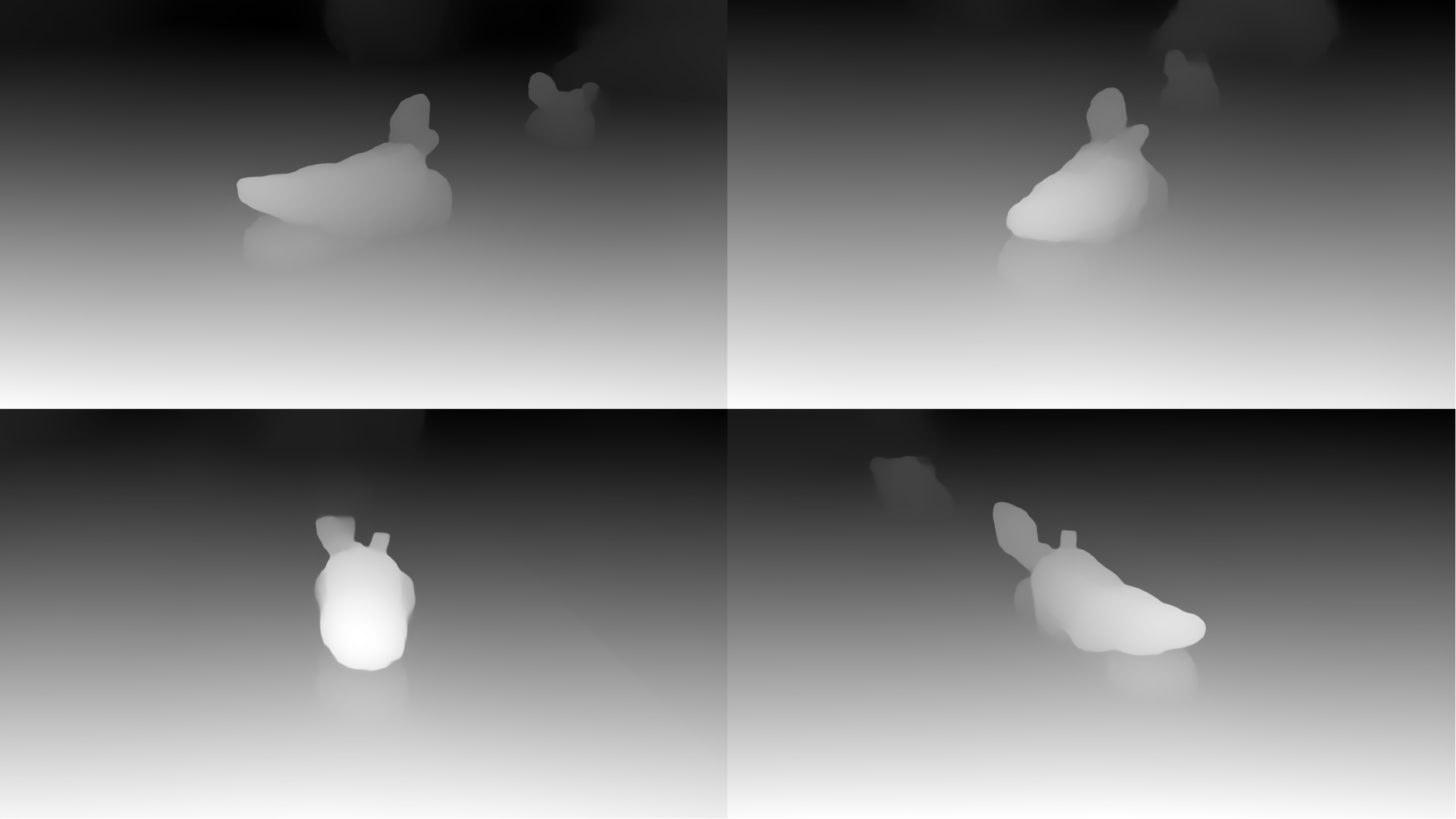}}
\caption{Predicted depth images using MiDaS corresponding to Fig. \ref{fig:nerf_predict}.}
\label{fig:nerf_depth}
\end{figure}

The second stage of the proposed pipeline is the prediction of depth images from novel view images predicted by the NeRF.
The monocular depth estimation model MiDaS is a zero-shot model (no fine-tuning is required), meaning that it can robustly perform depth prediction on a variety of data.
MiDaS was trained on ten datasets, and the network structure was based on RefineNet.
The depth image $d$ predicted by MiDaS, $\mathcal{N}_{MiDaS}$, can be expressed as follows:
\begin{equation}
d = \mathcal{N}_{MiDaS} \left\{ RGB \right\}.
\end{equation}
Figure \ref{fig:nerf_depth} presents the predicted depth images corresponding to Fig. \ref{fig:nerf_predict}

The final stage of the proposed pipeline is hologram prediction. In this study, ``Tensor Holography'' \cite{shi2021towards} was utilized. 
There are numerous implementations of deep learning-based hologram calculations. Among them, we adopted Tensor Holography because it is open-source and can predict high-quality RGB holograms with real-time frame rates.
The network structure of the Tensor Holography is based on ResNet. The input data is composed of an RGB image and the corresponding depth image, and the output is an RGB hologram. 
The predicted RGB hologams $H_{RGB}$ using Tensor Holography, denoted by $\mathcal{N}_{Hol}$ can be expressed as
\begin{equation}
H_{RGB} = \mathcal{N}_{Hol} \left\{ RGB, d \right\}.
\end{equation}
The stack number of convolutional layers in Tensor Holography is proportional to the spread of the point spread function of the light wave.

\begin{figure}[h]
\centering
\fbox{\includegraphics[width=\linewidth]{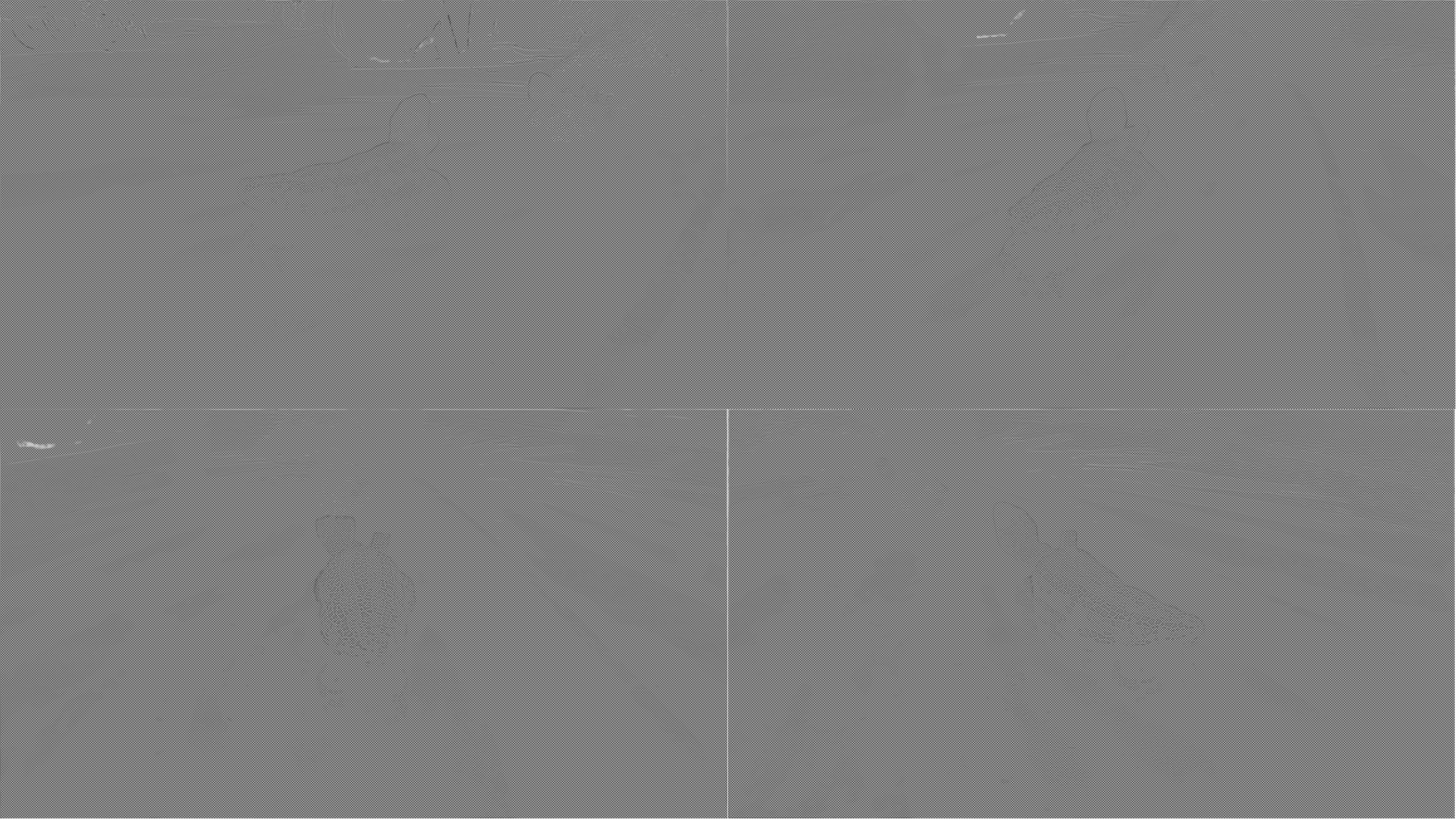}}
\caption{Predicted holograms from the proposed pipeline (only show green channel) corresponding to Figs. \ref{fig:nerf_predict} and \ref{fig:nerf_depth}. The holograms are encoded by a double-phase hologram technique.}
\label{fig:nerf_cgh}
\end{figure}

In summary, the entire proposed pipeline can be expressed as
\begin{equation}
H_{RGB} = \mathcal{N}_{Hol} \left\{ \mathcal{N}_{NeRF} \{{\bold v}\},~\mathcal{N}_{MiDaS} \{ \mathcal{N}_{NeRF} \{{\bold v}\} \}\right\}.
\end{equation}
Thus, during the prediction stage, the proposed pipeline requires only ${\bold v}$.
Figure \ref{fig:nerf_cgh} presents the predicted holograms (only the green channel) corresponding to Figs. \ref{fig:nerf_predict} and \ref{fig:nerf_depth} using the proposed pipeline. The output holograms are encoded using a double-phase hologram technique \cite{hsueh1978computer,sui2021band,shi2021towards}.
\add{Note that in the current pipeline, trained MiDaS and Tensor Holography have been used; only NeRF has been trained.}

\section{Results}
Subsequently, simulation and experimental results are presented.
The computational environment was as follows: an AMD Ryzen 5 4500 CPU (3.6 GHz) with 64 GB of memory and an Nvidia GeForce RTX 2080 SUPER graphics processing unit (GPU).
We prepared photographs of the park, as illustrated in Fig.\ref{fig:nerf_original} for verification.
The pixel pitch of the holograms was 8.0 $\mu$m, the distance between the 3D scene and the hologram was 0.6 mm, and the wavelengths of red, green, and blue were 650, 532, and 450 nm, respectively. The resolution of all the images was $1920 \times 1080$ pixels.
These parameters are fixed by the use of Tensor Holography.

\begin{figure*}[h]
\centering
\fbox{\includegraphics[width=14cm]{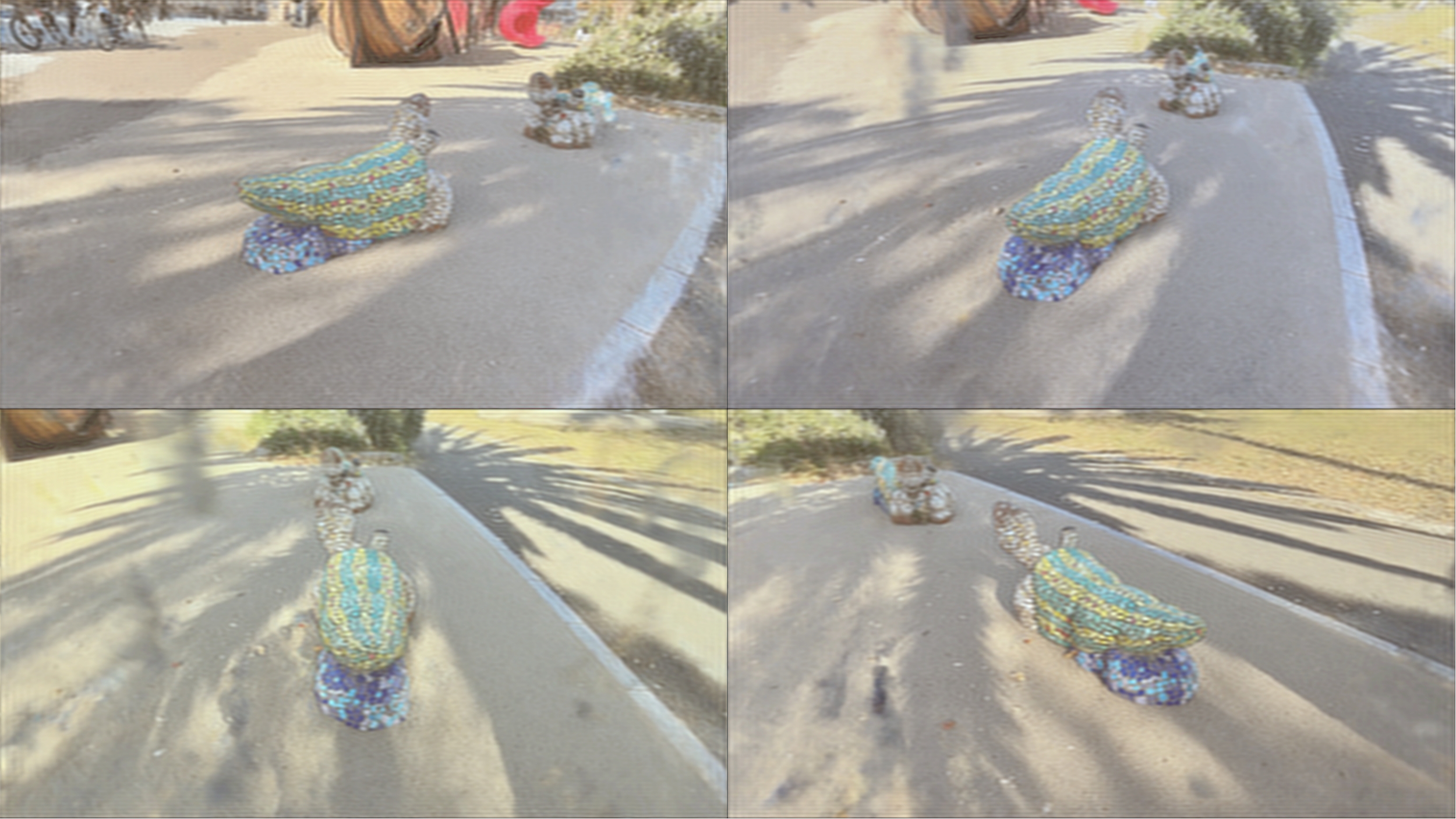}}
\caption{Simulated color reconstructions from the holograms of Fig. \ref{fig:nerf_cgh}.}
\label{fig:nerf_reconst}
\end{figure*}

\begin{figure}[h]
\centering
\fbox{\includegraphics[width=8cm]{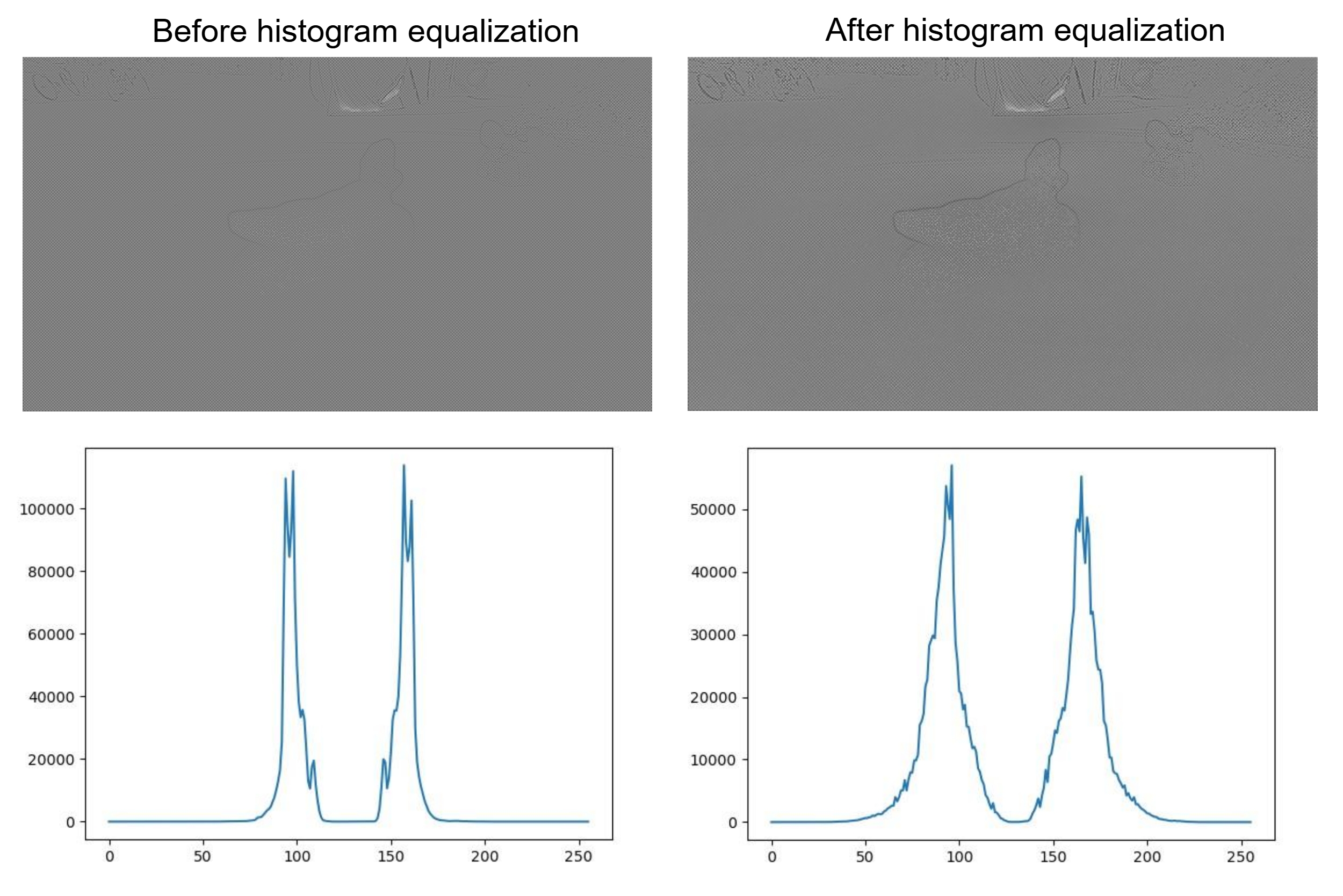}}
\caption{Holograms and histograms before/after the histogram equalization technique.}
\label{fig:histogram}
\end{figure}

\subsection{Simulations}
Figure \ref{fig:nerf_reconst} illustrates the simulated color reconstructions from the holograms predicted in Fig. \ref{fig:nerf_cgh}. The angular spectrum method \cite{matsushima2009band} was used for reconstruction.
The calculation times for each stage are presented in Fig. \ref{fig:outline} are discussed. 
The most time-consuming part was the optimization of the NeRF. We used 70 captured images, and the optimization time was approximately 1 min to achieve the image quality illustrated in Fig. \ref{fig:nerf_predict}. 
\add{The number of images to be input to NeRF is determined empirically from the quality of the output novel view images.}
Following optimization, the NeRF can predict new synthetic images in real-time.
The prediction times of MiDaS and Tensor holography per three color channels were about 5 seconds, respectively. Thus, the proposed pipeline can render holograms viewed from a new view position and an angle of approximately 10 seconds, except for the optimization time of the NeRF.
\add{With regard to Tensor Holography, this study used the TensorFlow model; it would be faster with TensorRT.}

The reconstructed images illustrated in Figs. \ref{fig:nerf_reconst} has low contrast compared to the original images in Fig. \ref{fig:nerf_original} due to a domain shift problem in Tensor Holography, which causes the different nature of the data used for training and the data to be predicted.
There are two methods to enhance contrast. The first involves transfer learning, wherein Tensor Holography is retrained using our target datasets. Although the retrained neural network predicts the contrast-corrected holograms, it may be necessary to retrain every new target dataset.

\begin{figure*}[h]
\centering
\fbox{\includegraphics[width=\linewidth]{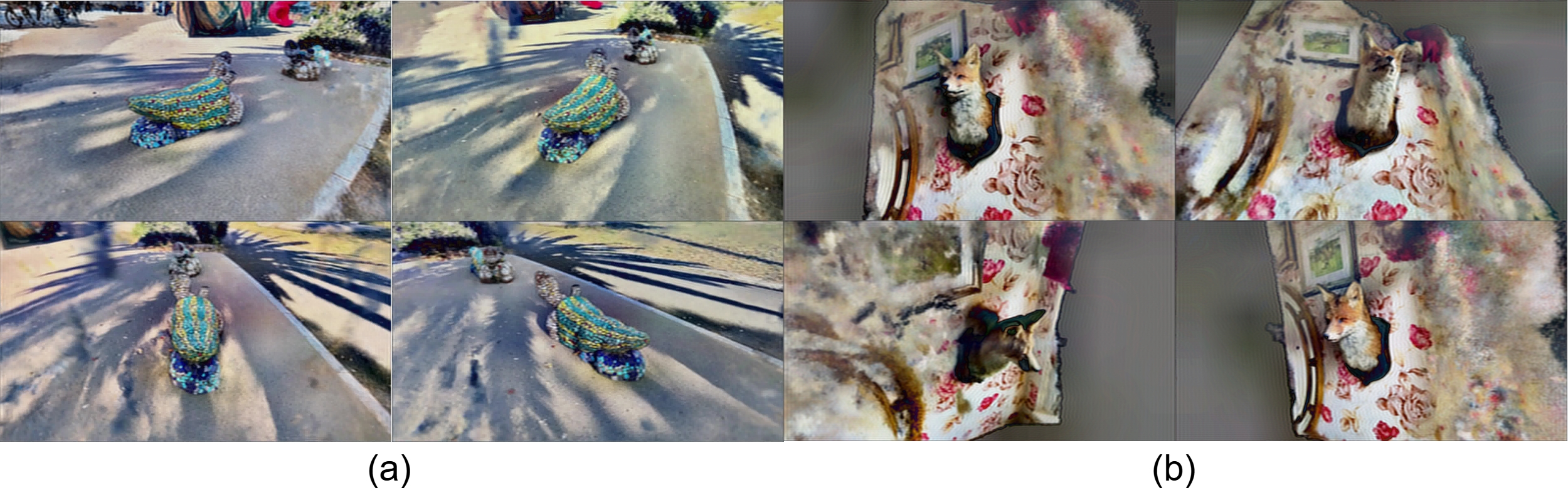}}
\caption{Simulated reconstructions after applying CLAHE to holograms predicted 
from Tensor Holography: (a) corresponding to Fig. \ref{fig:after_clahe}(a) and  (b) corresponding to Fig. \ref{fig:after_clahe}(b) (Visualizations 1 and 2).}
\label{fig:after_clahe}
\end{figure*}

The second method uses classical contrast enhancement algorithms.
\add{Denoting a hologram as $h$, a propagation operator as $\mathcal{P}$, the complex amplitude of the reproduced image can be expressed as $u=\mathcal{P}(h)$.
Defining a contrast enhancement as $\mathcal{E}$, the complex amplitude of the reproduced image can be expressed as $\mathcal{E}u = \mathcal{P}(\mathcal{E}h)$, and we can observe a contrast-enhanced reproduced image of $|\mathcal{E}u|^2$.}
Among numerous contrast enhancement algorithms, 
contrast-limited adaptive histogram equalization (CLAHE) \cite{pizer1987adaptive} was applied to the holograms predicted from Tensor Holography.
In CLAHE, a hologram is divided into nonoverlapping regions, and conventional contrast enhancement is applied to small regions. 
Figure \ref{fig:histogram} presents the holograms and histograms of Fig. \ref{fig:nerf_cgh} before and after applying CLAHE. 
The simulated reconstruction of the holograms after applying CLAHE is illustrated in Fig. \ref{fig:after_clahe}(a).
Thus, the contrast of the reproduced images can be enhanced.



Figure \ref{fig:after_clahe}(b) shows reproduced images from another hologram using the proposed pipeline. Unlike Fig. \ref{fig:nerf_original}, this original scene was computer-graphically rendered and new viewpoint holograms were generated from the rendered images.

 \add{Note that artifacts in the air are not from depth estimation errors, but from NeRF estimation errors. More images need to be taken for more accurate estimation.}

\subsection{Optical experiments}
\begin{figure}[h]
\centering
\fbox{\includegraphics[width=8cm]{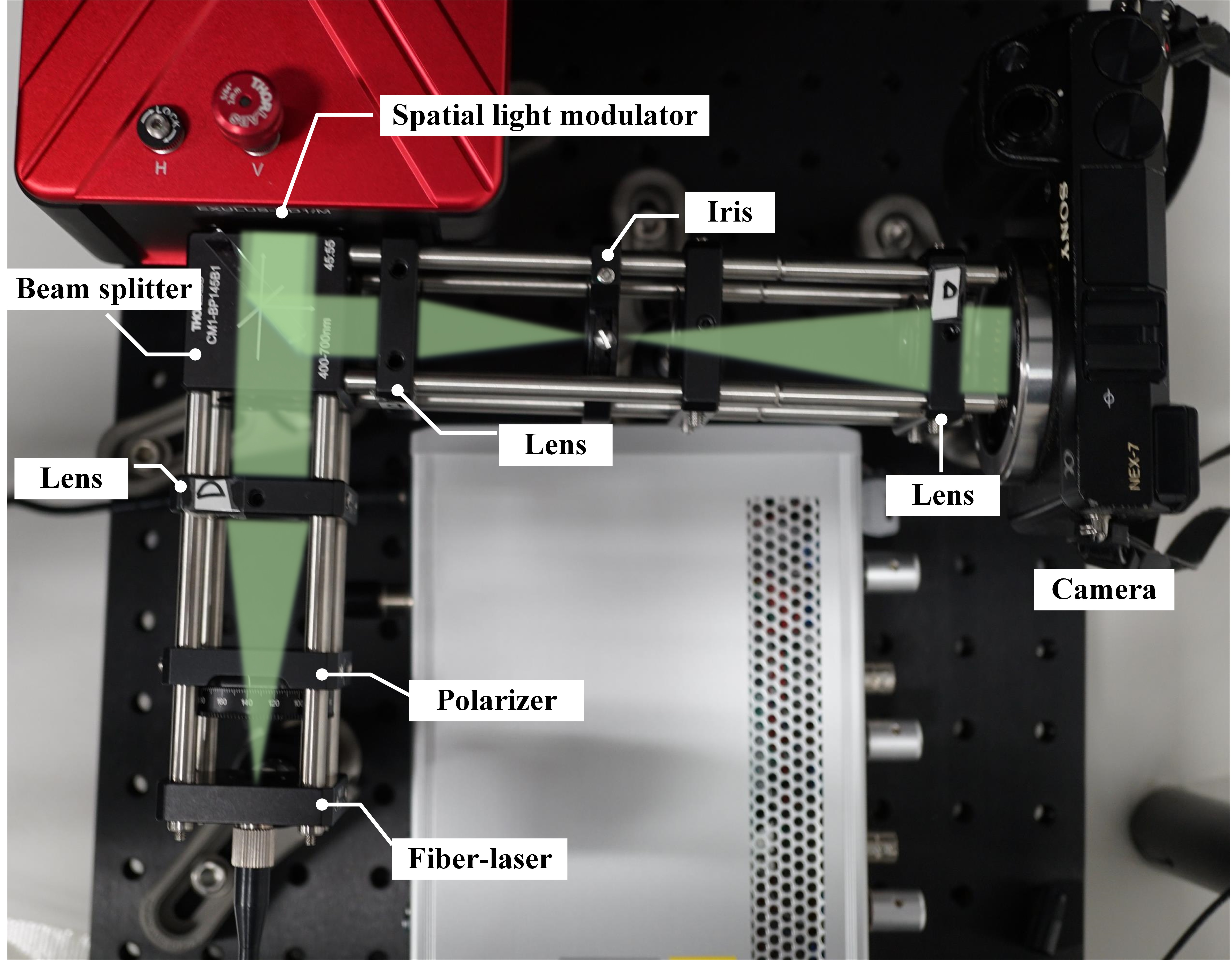}}
\caption{Optical setup.}
\label{fig:optical_setup}
\end{figure}

\begin{figure}[ht]
\centering
\fbox{\includegraphics[width=\linewidth]{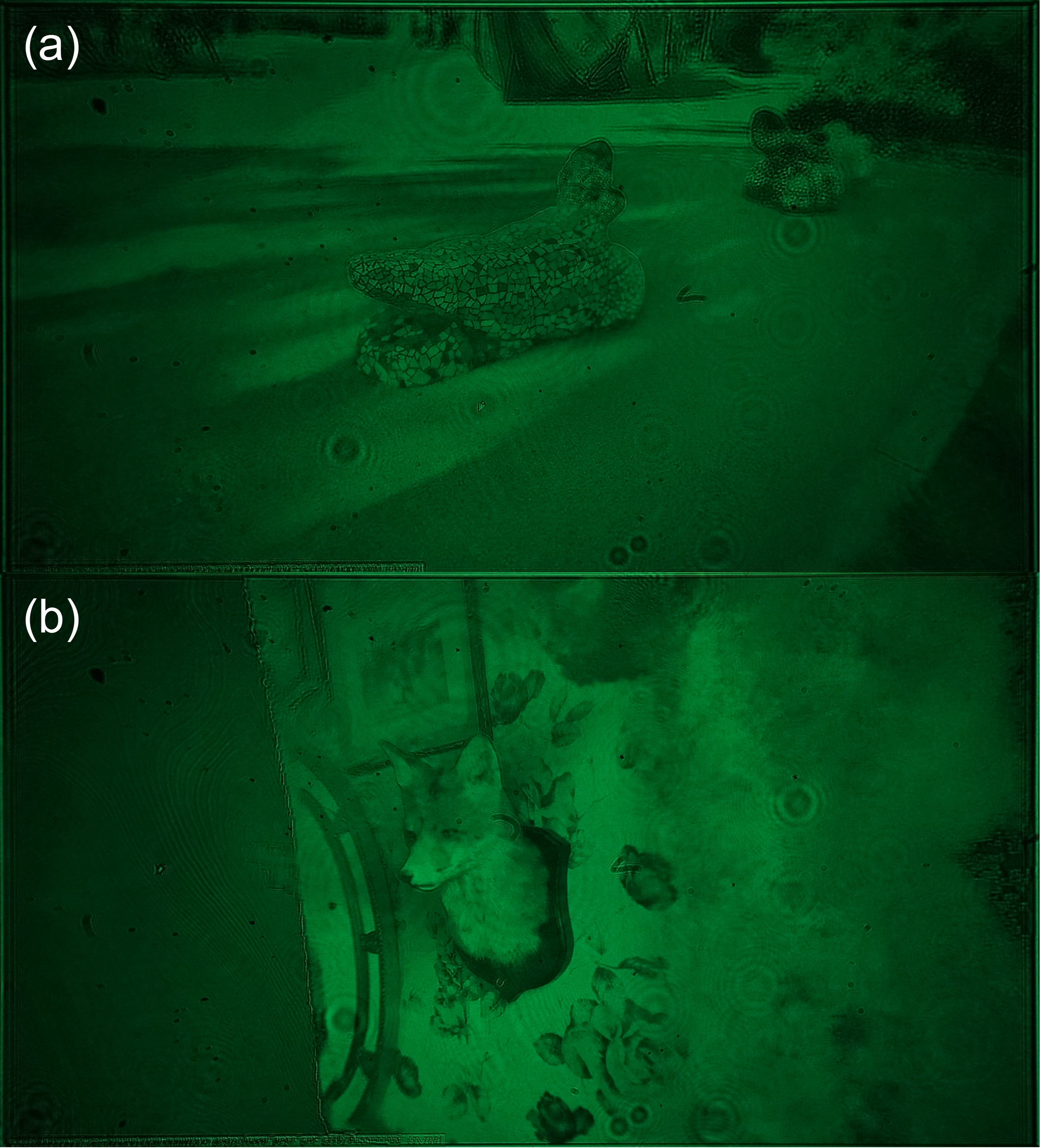}}
\caption{Optical reconstructions: (a) reconstruction corresponding to Fig. \ref{fig:after_clahe}(a) (visualization 3), and (b) reconstruction corresponding to Fig. \ref{fig:after_clahe}(b) (visualization 4).}
\label{fig:optical_reconst}
\end{figure}

Optical experiments were conducted as illustrated in Fig. \ref{fig:optical_setup}.
The optical system consisted of a single-mode fiber laser source, lenses, a polarizer, a 4f filter, a spatial light modulator, and a camera. 
A linear phase pattern was added to the holograms to avoid overlapping of the signal and direct light.
The wavelength was set at 520 nm. 
Figure \ref{fig:optical_reconst} illustrates the optical reconstruction: (a) reconstruction corresponding to Fig. \ref{fig:after_clahe}(a), and (b) reconstruction corresponding to Fig. \ref{fig:after_clahe}(b).

\section{Discussions}
The effectiveness of the proposed method, the current problems, and ideas for future improvements are discussed.
NeRF allows the proposed pipeline to predict the hologram in any viewing direction.
Considering this feature, the proposed pipeline can be extended to generate cylindrical and spherical holograms \cite{yamaguchi2008fast,tachiki2006fast} and light-field-based holograms \cite{yamaguchi2016light}. 
By arranging cylindrical or spherical holographic media ( SLMs), we can observe 3D images at $360^\circ$ or omnidirectional positions. 
This is illustrated in Figs. \ref{fig:nerf_reconst} and \ref{fig:optical_reconst} that holographic reconstructions can be obtained in arbitrary view directions.
For light-field-based hologram calculations, a camera in the NeRF translates horizontally and vertically to produce multiple images.
Recently, deep neural networks have been proposed that predict 3D holograms using only RGB images without depth images \cite{chang2023picture,ishii2023multi}. 
If these neural networks are incorporated into the proposed pipeline instead of Tensor Holography, the pipeline will be more straightforward because a depth prediction network, such as MiDaS, can be removed. 

Although the effectiveness of the proposed pipeline was demonstrated, the current pipeline has several limitations. Instant NeRF is a high-speed version of NeRF, however, it requires a relatively long optimization time. In addition, multiple captured images were required. More recently, NeRF techniques have significantly improved optimization, prediction times, and image quality.
For instance, pixelNeRF \cite{yu2021pixelnerf} can optimize NeRF using only a single or a few images, which relaxes the cumbersome process of taking photos and improves the speed of optimization.
Another problem is the generation of holograms using deep learning.
\add{Although Tensor Holography can predict high-quality holograms in real-time, it cannot easily represent deep 3D scenes because it does not render point-spread functions with high-frequency components.} To enhance the depth range, the number of ResNet modules in Tensor Holography should be increased, leading to an increase in the computational burden.
Another candidate for solving this problem is the Fourier basis-inspired technique for predicting holograms using high-frequency signals \cite{zhu2023computer}.  

\section{Conclusions}
This study proposes a hologram generation pipeline composed of a 3D reconstruction neural network (NeRF), depth prediction of MiDaS, and a hologram generator of Tensor Holography. 
Following optimization, the proposed pipeline can predict RGB holograms viewed in arbitrary directions within a reasonable time using only the view vector ${\bold v}$.
The current pipeline has problems in that it requires a relatively long optimization time and many images for NeRF. In addition, Tensor Holography is limited to shallow-depth representations. However, the use of an improved hologram predictor and the NeRF can solve these problems.
The depth predictor can be removed by using neural networks \cite{chang2023picture,ishii2023multi}, resulting in a simpler network structure and faster hologram predictions.

\begin{backmatter}
\bmsection{Funding}
Japan Society for the Promotion Science (22H03607, 19H01097); IAAR Research Support Program, Chiba University, Japan.

\bmsection{Disclosures}The authors declare no conflicts of interest.

\bmsection{Data Availability Statement}
Data underlying the results presented in this paper are not publicly available at this time but may be obtained from the authors upon reasonable request.

\end{backmatter}

\bibliography{paper}

\bibliographyfullrefs{sample}



\end{document}